\documentclass[11pt,a4paper]{article}
\usepackage[hyperref]{emnlp2020}
\usepackage{times}
\usepackage{textcomp}
\usepackage{latexsym}
\usepackage{amssymb,mathtools}
\usepackage{graphicx,paralist,multirow}
\renewcommand{\textrightarrow}{$\rightarrow$}
\renewcommand{\textleftarrow}{$\leftarrow$}

\DeclareMathOperator{\R}{\mathbb{R}}

\usepackage{microtype}

\aclfinalcopy 

\title{Incorporating Commonsense Knowledge Graph in Pretrained Models for Social Commonsense Tasks}

\author{Ting-Yun Chang$^1$,  Yang Liu$^2$, Karthik Gopalakrishnan$^2$, Behnam Hedayatnia$^2$,\\ 
\bf{Pei Zhou$^3$, Dilek Hakkani-T\"{u}r$^2$}  \thanks{$*$ Work was done while Ting-Yun Chang and Pei Zhou were interns at Amazon.}  \\
$^1$ Academia Sinica, Taiwan; \\
$^2$ Alexa AI, Amazon, USA; \\
$^3$ USC, USA \\
  \texttt{r06922168@ntu.edu.tw, \{yangliud,karthgop,behnam,hakkanit\}@amazon.com}}

\date{}

\begin{document}
\maketitle
\begin{abstract}
Pretrained language models have excelled at many NLP tasks recently; however, their social intelligence is still unsatisfactory. To enable this, machines need to have a more general understanding of our complicated world and develop the ability to perform commonsense reasoning besides fitting the specific downstream tasks. External commonsense knowledge graphs (KGs), such as ConceptNet, provide rich information about words and their relationships. Thus, towards general commonsense learning, we propose two approaches to \emph{implicitly} and \emph{explicitly} infuse such KGs into pretrained language models. We demonstrate our proposed methods perform well on SocialIQA, a social commonsense reasoning task, in both limited and full training data regimes.
\end{abstract}

\section{Introduction}
Empowering machines with commonsense has become a hot topic recently. 
Past research efforts for this problem include the construction of various data sets and models.  
Several commonsense data sets have been commonly used in past work to develop machines' commonsense capability~\cite{talmor2019commonsenseqa,huang2019cosmos,zellers-etal-2019-hellaswa,sap2019socialiqa,sakaguchi2019winogrande,gordon2012semeval,rajani2019explain}.
In particular, SocialIQA~\cite{sap2019socialiqa} is a multiple-choice QA data set for probing machine's emotional and social intelligence in a variety of everyday situations, which is the data set used in this study. 
To improve the modeling approaches for the SocialIQA and other commonsense tasks, 
\citet{shwartz2020unsupervised} and \citet{bosselut2019dynamic} focused on zero-shot setting using pretrained language models. 
\citet{khashabi2020unifiedqa} reformulated the multi-choice setup used in most data sets  as a generation task and achieved impressive performance by fine-tuning T5~\cite{raffel2019exploring}. 
Recently there is an increasing effort to utilize external knowledge bases to incorporate commonsense information underlying the text \cite{shwartz2020unsupervised,mitraadditional,ji2020generating,ji2020language}. 

While most prior work on SocialIQA utilized large pretrained language models~\cite{devlin2019bert,liu2019roberta,radford2018improving, radford2019language,raffel2019exploring}, we argue that such a challenging task requires commonsense reasoning of social events, and simply fine-tuning the model to fit the task is insufficient. 
We believe it would be beneficial if the model can learn from knowledge-rich resources such as ConceptNet~\cite{liu2004conceptnet}, and thus have a broader and deeper understanding of the information not present in the provided context and answer candidates.

In this paper, we propose two approaches tailored to large pretrained language models to utilize existing knowledge graph (KGs) for downstream commonsense tasks. The first approach leverages the KGs \emph{implicitly} by pretraining on the relevant tuples to the SocialIQA task, while the second one maintains a dynamic knowledge base during fine-tuning, utilizing KGs \emph{explicitly} via an attention mechanism. 
Our experiments demonstrate the effectiveness of both approaches on SocialIQA under limited and full training data regimes, and the critical role of relevant knowledge. 

\section{Problem Formulation and Baseline}
\label{sec:problem_form}
\begin{figure}[ht]
\centering
\includegraphics[width=7.5cm]{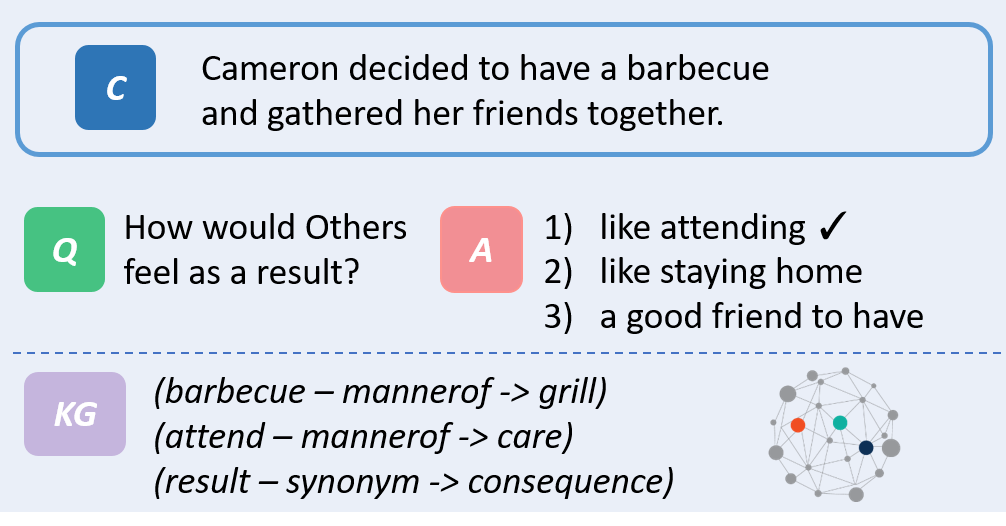}
\caption{An instance in SocialIQA and our retrieved tuples from ConceptNet.}
\label{fig:example}
\end{figure}

In SocialIQA, given a context $C$ of an event and a corresponding question $Q$, the goal is to select the correct choice from the answer set $A=(A_1, A_2, A_3)$. An example is shown in Figure~\ref{fig:example}, with the provided context, question, and three answer candidates.

A typical approach (as used in ~\citet{sap2019socialiqa}) for solving this kind of multiple-choice problems with a pretrained Transformer-based language model is by concatenating $C$, $Q$, and $A_i$ with a separator token, and then letting the model output a score via a multi-layer perceptron (MLP) built on top of the final hidden representation of the classifier token $[CLS]$. Finally, scores for each data point are normalized across all $(C, Q, A_i)$ instances with softmax, and cross-entropy loss is applied for model training.

Since RoBERTa~\cite{liu2019roberta} has shown competitive performance on the SocialIQA task, we use it as a baseline model in this study. Furthermore, in addition to multiple-choice classification, we perform masked language modeling (MLM)~\cite{devlin2019bert}, masking $15\%$ tokens in the concatenation of $C, Q$, and the correct answer $A_{y^*}$, when fine-tuning on the SocialIQA task.

\section{Incorporating Commonsense Knowledge Graph}
In this section, we introduce two methods to incorporate a given KG into our pretrained model. We experiment with both ATOMIC~\cite{sap2019atomic} and ConceptNet~\cite{liu2004conceptnet} as our KGs. 
\begin{itemize}
    \item \textbf{ATOMIC} focuses on inferential knowledge of everyday situations. Each node in ATOMIC is a social event, containing 9 \emph{if-then} relation types. Note that though SocialIQA is derived from ATOMIC, it has been rewritten by crowd workers~\cite{sap2019socialiqa}.
    \item \textbf{ConceptNet} represents general words and phrases that people use and the commonsense relationships between them, such as \emph{IsA, AtLocation, Desires, Synonym}.
\end{itemize} 


\subsection{Querying Knowledge Graph}
For both methods, we first extract keywords in the input data to query ConceptNet using only the lemmatized \emph{noun, verb, adjective} words in $(C, Q, A)$ as queries, with stop words excluded.
First, we find the node corresponding to each query in the KG, and retrieve all the connected tuples \emph{(query, relation with weight, tail)} within one hop. We then sort all the retrieved tuples by their $relation\ weight \times query's \ idf$, and keep the top-$k$ tuples for each data point.\footnote{Initially, we tried to extract the shortest path between keywords in the KG similar to~\citet{shwartz2020unsupervised}. However, as ConceptNet does not disambiguate word senses, we observed that such paths usually deviate from the original semantics. For example, consider \emph{``C: Cameron decided to have a \underline{barbecue} and gathered her friends together. $A_1$: like \underline{attending}."} The path we found between \emph{barbecue} and \emph{attend} is: \emph{barbecue--isa\textrightarrow dish--synonym\textrightarrow serve\textleftarrow synonym--attend}. Similarly, since we found that some of the retrieved tuples within one hop are already irrelevant, we did not use more hops to retrieve relevant tuples.}
In Figure~\ref{fig:example}, the bottom shows examples of retrieved tuples from ConceptNet for an instance in the SocialIQA data. 
Note that since ATOMIC highly overlaps with SocialIQA, we do not extract keywords to query the KG but pretrain the model on the entire KG.

\subsection{Pretraining Language Models on Retrieved Concepts}
In the first approach, we leverage the KG via infusing it into the pretraining step.  Using the SocialIQA data as queries, we first retrieve tuples from the KGs as described above, and then convert them to textual forms. 
To enable this conversion, we hand-crafted templates for different relations. For example, a tuple in ConceptNet \emph{(barbecue, hascontext, cooking)} would be converted into ``\emph{barbecue is a word used in the context of cooking.}" 
When using ATOMIC, because there are some blanks and unknown names such as \emph{ "PersonX meets \_ for lunch}", we replace \emph{PersonX} and \emph{PersonY} with two different common \emph{last names} to avoid gender bias, and following~\citet{mitraadditional}, we utilize the pretrained BERT-large's MLM head to fill in the blanks.
After these steps, we build a corpus derived from concepts in the KGs relevant to the SocialIQA task. 

We then train our RoBERTa-based models using such a corpus with the MLM loss~\cite{devlin2019bert}, masking either the head or tail entities, e.g., \emph{barbecue} or \emph{cooking}.
Further training the pretrained models on such a corpus is expected to implicitly learn commonsense knowledge in the KGs that is relevant to SocialIQA. 
Finally, we continue to fine-tune the model on the SocialIQA task, similar to the baseline described in Section~\ref{sec:problem_form}.

\subsection{Modeling Concepts Via Attention}
In the second approach, we treat the retrieved tuples as items in a cached external knowledge base (KB), which dynamically changes based on every input instance. The model can then decide the importance of each item and leverage them accordingly.

\paragraph{KG Attentive Representations}
\label{section:variants}
Motivated by previous work on question answering~\cite{Seo2017Bidirectional, zhu2018hierarchical, Wang2018YuanfudaoAS, huang2019cosmos}, which uses attention among different segments of the input, here we treat the knowledge tuples as a new segment. Specifically, we concatenate the top-$k$ retrieved tuples and map them into the space of RoBERTa's final hidden representations as an additional segment, and then attend to it using RoBERTa's last hidden representation to generate a new KG-attentive sentence representation. 

Formally, let $d$ be the hidden dimension, $l$ be the sequence length of the input, $H_R\in {\R}^{l\times d}$ be RoBERTa's final hidden representation for the SocialIQA input sequence for a given candidate, and $H_{KG}\in {\R}^{k\times d}$ is the representation of the $k$ encoded tuples. We attend to $H_{KG}$ from $H_R$:
\begin{align}
\hat{H_R} &= H_R\ W_1 + \mathbf{1}*b_1^T,\nonumber \\
\hat{H}_{KG} &= H_{KG}\ W_1 + \mathbf{1}*b_1^T, \nonumber \\
H_R^{KG} &= Softmax(\frac{\hat{H_R} \hat{H}_{KG}^T}{\sqrt{d}}) \ H_{KG}
\label{eq:attn}
\end{align}
where $W_1 \in {\R}^{d\times d}$, $\mathbf{1} \in {\R}^{l}$ (a vector of all-ones), $b_1 \in {\R}^{d}$, and $H_R^{KG}\in {\R}^{l\times d}$ is the KG-attentive sentence representation.

\paragraph{Encoding Knowledge Tuples}
To obtain $H_{KG}$, we need to represent the tuples and project them to the RoBERTa's hidden representation space. We first convert the tuples into fixed embeddings with three different approaches: 
\begin{itemize}
\item Pretrained KG embeddings based on ConceptNet via TransE~\cite{bordes2013translating, zhou2018commonsense}.
\item Pretrained word embeddings retrofitted by ConceptNet~\cite{speer2017conceptnet}, where its training adjusts a word's embeddings to be close to those of its neighbors in the graph.
\item Encoded tuple-converted text with templates and pretrained universal sentence encoder (USE)~\cite{cer2018universal}, a Transformer-based sentence encoder that transforms text into vectors that can be used for text classification and semantic similarity.
\end{itemize}

Then we transform these embeddings of the top-$k$ tuples using a linear transformation that is learned during training, and then concatenate all of them to form the knowledge representation $H_{KG} \in {\R}^{k\times d}$. 

\begin{figure}[ht]
\centering
\includegraphics[width=7.5cm]{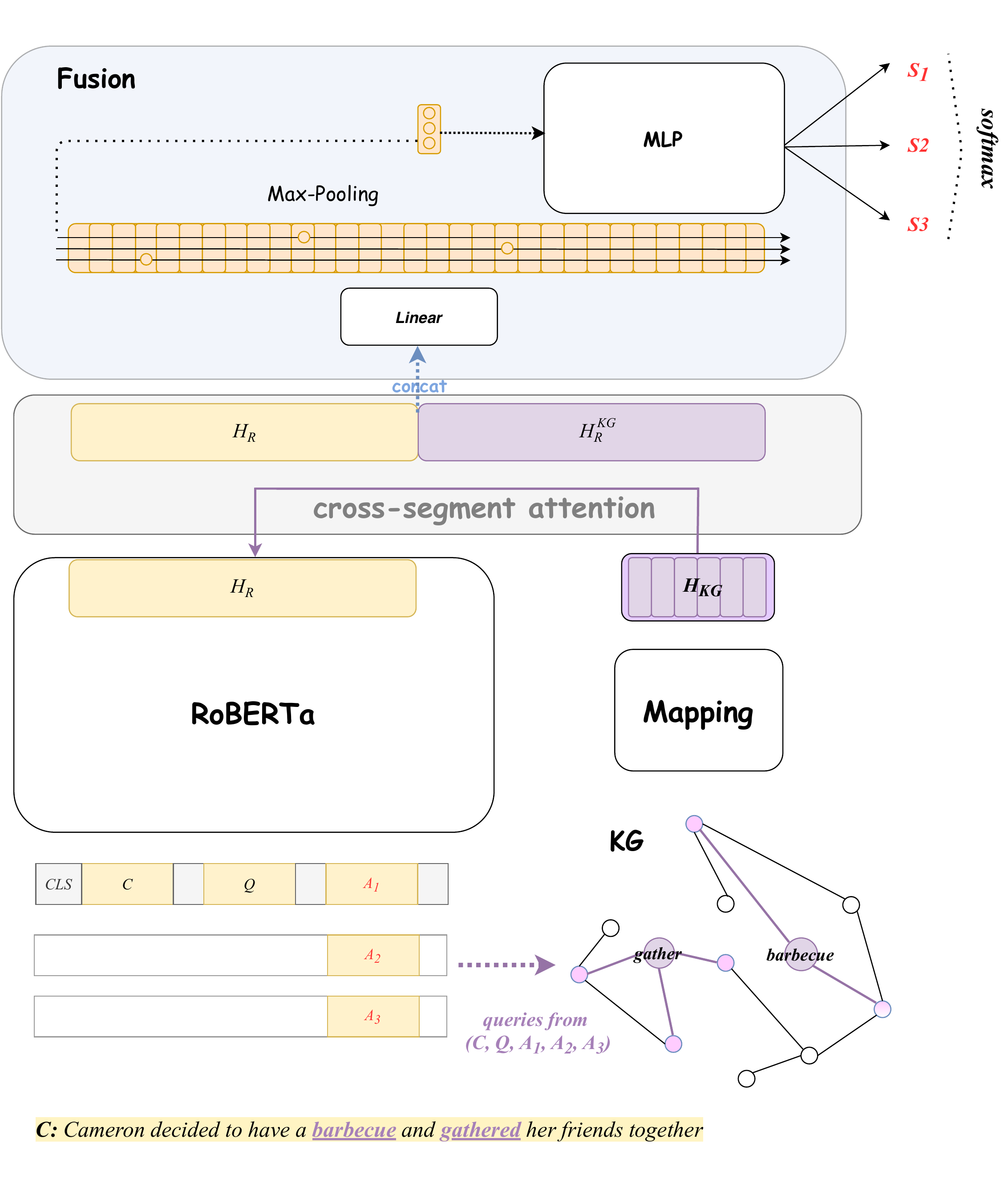}
\caption{Illustration of the proposed model incorporating external KGs for SocialIQA.}
\label{fig:KG}
\end{figure}

\paragraph{Fusion Layer}
We then combine $H_R$ and $H_R^{KG}$ with a \emph{fusion layer}. Formally,
\begin{align}
\Tilde{H_R} &= [H_R \oplus H_R^{KG}]\ W_2 + \mathbf{1}*b_2^T,\nonumber \\
\Tilde{h_R} &= max\{\Tilde{H_R}\},\nonumber \\
S &= MLP(\Tilde{h_R}),
\label{eq:attn}
\end{align}
where we first transform the concatenation (denoted by $\oplus$) of $H_R$ and $H_R^{KG}$ to $\Tilde{H_R} \in {\R}^{l\times d}$, and then perform max-pooling along the sequence dimension to obtain the condensed representation $\Tilde{h_R} \in {d}$ for classification. Finally, we get the score $S \in {\R}^1$ for each answer via a multilayer perceptron (MLP). The model architecture is illustrated in Figure~\ref{fig:KG}.

\section{Experiments and Results}
\subsection{Experimental Setup}
We use Hugging Face's \texttt{transformers} toolkit\footnote{\url{https://huggingface.co/transformers/}} and train our models on the 33k SocialIQA training instances, running hyper-parameters search over the learning rate in $\{5e-6,\ 1e-5,\ 2.5e-5\}$, and the ~\emph{effective batch size} (number of GPUs $\times$ batch size per GPU $\times$ gradient accumulation steps) in $\{8,\ 16,\ 32\}$ for the proposed models and baselines respectively, and report their best performance on the dev set. We set the maximum returned tuples of each instance to $k=30$. 

\subsection{Results}
Table~\ref{table:proposed} shows the results on the dev set using different methods.
We can see the first pretraining method using ATOMIC performs well, which is not surprising since it is the partial source of SocialIQA, and it is likely that the model may have seen related information about dev/test set during pretraining.  
On the other hand, the performance of ConceptNet-pretrain suggests that without a sophisticated design, exposing too many irrelevant tuples from ConceptNet in  pretraining  may compromise the model's performance on the downstream task. This finding is consistent with ~\citet{gururangan2020don}. In our analysis, we did find that some of the extracted tuples are noisy, mainly because ConceptNet is comprehensive, but it does not contain annotations for different word senses.

\begin{table}[ht]
  \centering
  \begin{tabular}{|l|c|}
  \hline
	\bf Model & \bf Accuracy (\%) \\
  \hline\hline
    Baseline \citet{sap2019socialiqa} & 78.0 \\
    \hline
     ConceptNet-pretrain & 76.8 \\
    ATOMIC-pretrain & \bf{79.1} \\ \hline
    ConceptNet-attention-TransE & 78.5 \\
    ConceptNet-attention-Retrofit & 78.7 \\
    ConceptNet-attention-USE & \bf{79.2} \\

  \hline
  \end{tabular}
  \caption{Comparison of different models on the SocialIQA dev set.}
  \label{table:proposed}
\end{table}

The second method, however, seems promising. Among its three variants (see Section~\ref{section:variants}), the one using TransE as knowledge embeddings performed the worst, possibly because of its much smaller dimension of the pretrained TransE embeddings ($\in {\R}^{100}$) we adopted\footnote{\url{http://coai.cs.tsinghua.edu.cn/hml/dataset/\#commonsense}}, compared to the other two variants ($300$ and $512$) and RoBERTa-large's hidden dimension $d=1024$. These results indicate that the second method of using ConceptNet is less sensitive to the noisy tuples because of the explicit attention mechanism, which allows the model to utilize the items in the KG selectively.

\subsection{Few-Shot Learning} 
To demonstrate the effective utilization of external KGs, we now investigate performance in the limited training data regime.
We fine-tuned our model on $5\%, 10\%,$ and $20\%$ of SocialIQA's training instances, respectively. We compare both the explicit method, ConceptNet-attention-USE, and the implicit method, ATOMIC-pretrain, with the typical implementation of RoBERTa as the baseline. We apply MLM on SocialIQA fine-tuning (Section~\ref{sec:problem_form}) in all the three models, since we have found it helps stabilize the training.

The results in Table~\ref{tab:few_shot} show that ATOMIC-pretrain performs especially well, even though it only relies on the pretraining phase to infuse the ATOMIC graph, reaching $72.9\%$ when only $5\%$ of training instances are used. ConceptNet-attention-USE performs the worst on the $5\%$ setting, but better than the baseline on the other two settings. Note that a BERT-base model trained on the full training set only achieves $63.3\%$~\cite{sap2019socialiqa}, showing that RoBERTa model may already learn some commonsense in its pretraining phase. 
Furthermore, our proposed methods demonstrate that with the external knowledge graphs on relevant domains, we can obtain even better results when only a small number of annotated training instances for the downstream task are available. 

\begin{table}[ht]
    \centering
    \begin{tabular}{|c|c|c|c|} \hline
 & 5\% & 10\% & 20\% \\ \hline
RoBERTa + MLM  &    70.3   & 72.3 & 73.0  \\
ATOMIC-pretrain & 72.9 & 73.3 & 76.0 \\ 
ConceptNet-attn-USE & 69.7 & 73.3 & 74.6 \\ \hline
    \end{tabular}
    \caption{Results (accuracy \%) when using few training instances for model fine-tuning.}
    \label{tab:few_shot}
\end{table}

\section{Conclusion}
In this paper, we propose two methods to introduce KGs into  pretrained language models for commonsense tasks. The first one implicitly infuses relevant knowledge into  MLM pretraining, while the second method uses the attention mechanism to allow pretrained language models to explicitly utilize the dynamic query tuples. Our experiments on the SocialIQA task show that leveraging external KGs via attention outperforms the baseline pretrained language models, and the quality of the relevant graphs matters for downstream task performance. 
Our work can be further improved by designing better algorithms for KG retrieval in the future.
Although our experiments have focused on SocialIQA with ConceptNet and ATOMIC, our method can be generalized to other similar tasks to leverage knowledge graphs.

\bibliographystyle{acl_natbib}
\bibliography{anthology,emnlp2020}

\begin{thebibliography}{28}
\expandafter\ifx\csname natexlab\endcsname\relax\def\natexlab#1{#1}\fi

\bibitem[{Bordes et~al.(2013)Bordes, Usunier, Garcia-Duran, Weston, and
  Yakhnenko}]{bordes2013translating}
Antoine Bordes, Nicolas Usunier, Alberto Garcia-Duran, Jason Weston, and Oksana
  Yakhnenko. 2013.
\newblock Translating embeddings for modeling multi-relational data.
\newblock In \emph{Advances in neural information processing systems}, pages
  2787--2795.

\bibitem[{Bosselut and Choi(2019)}]{bosselut2019dynamic}
Antoine Bosselut and Yejin Choi. 2019.
\newblock Dynamic knowledge graph construction for zero-shot commonsense
  question answering.
\newblock \emph{arXiv preprint arXiv:1911.03876}.

\bibitem[{Cer et~al.(2018)Cer, Yang, Kong, Hua, Limtiaco, John, Constant,
  Guajardo-Cespedes, Yuan, Tar et~al.}]{cer2018universal}
Daniel Cer, Yinfei Yang, Sheng-yi Kong, Nan Hua, Nicole Limtiaco, Rhomni~St
  John, Noah Constant, Mario Guajardo-Cespedes, Steve Yuan, Chris Tar, et~al.
  2018.
\newblock Universal sentence encoder for english.
\newblock In \emph{Proceedings of the 2018 Conference on Empirical Methods in
  Natural Language Processing: System Demonstrations}, pages 169--174.

\bibitem[{Devlin et~al.(2019)Devlin, Chang, Lee, and
  Toutanova}]{devlin2019bert}
Jacob Devlin, Ming-Wei Chang, Kenton Lee, and Kristina Toutanova. 2019.
\newblock {BERT}: Pre-training of deep bidirectional transformers for language
  understanding.
\newblock In \emph{Proceedings of the 2019 Conference of the North {A}merican
  Chapter of the Association for Computational Linguistics (NAACL)}, pages
  4171--4186.

\bibitem[{Gordon et~al.(2012)Gordon, Kozareva, and
  Roemmele}]{gordon2012semeval}
Andrew Gordon, Zornitsa Kozareva, and Melissa Roemmele. 2012.
\newblock Semeval-2012 task 7: Choice of plausible alternatives: An evaluation
  of commonsense causal reasoning.
\newblock In \emph{* SEM 2012: The First Joint Conference on Lexical and
  Computational Semantics--Volume 1: Proceedings of the main conference and the
  shared task, and Volume 2: Proceedings of the Sixth International Workshop on
  Semantic Evaluation (SemEval 2012)}, pages 394--398.

\bibitem[{Gururangan et~al.(2020)Gururangan, Marasovi{\'c}, Swayamdipta, Lo,
  Beltagy, Downey, and Smith}]{gururangan2020don}
Suchin Gururangan, Ana Marasovi{\'c}, Swabha Swayamdipta, Kyle Lo, Iz~Beltagy,
  Doug Downey, and Noah~A Smith. 2020.
\newblock Don't stop pretraining: Adapt language models to domains and tasks.
\newblock \emph{arXiv preprint arXiv:2004.10964}.

\bibitem[{Huang et~al.(2019)Huang, Le~Bras, Bhagavatula, and
  Choi}]{huang2019cosmos}
Lifu Huang, Ronan Le~Bras, Chandra Bhagavatula, and Yejin Choi. 2019.
\newblock Cosmos qa: Machine reading comprehension with contextual commonsense
  reasoning.
\newblock In \emph{Proceedings of the 2019 Conference on Empirical Methods in
  Natural Language Processing and the 9th International Joint Conference on
  Natural Language Processing (EMNLP-IJCNLP)}, pages 2391--2401.

\bibitem[{Ji et~al.(2020{\natexlab{a}})Ji, Ke, Huang, Wei, and
  Huang}]{ji2020generating}
Haozhe Ji, Pei Ke, Shaohan Huang, Furu Wei, and Minlie Huang.
  2020{\natexlab{a}}.
\newblock Generating commonsense explanation by extracting bridge concepts from
  reasoning paths.
\newblock \emph{AACL-IJCNLP}.

\bibitem[{Ji et~al.(2020{\natexlab{b}})Ji, Ke, Huang, Wei, Zhu, and
  Huang}]{ji2020language}
Haozhe Ji, Pei Ke, Shaohan Huang, Furu Wei, Xiaoyan Zhu, and Minlie Huang.
  2020{\natexlab{b}}.
\newblock Language generation with multi-hop reasoning on commonsense knowledge
  graph.
\newblock In \emph{Proceedings of the 2020 Conference on Empirical Methods in
  Natural Language Processing (EMNLP)}.

\bibitem[{Khashabi et~al.(2020)Khashabi, Khot, Sabharwal, Tafjord, Clark, and
  Hajishirzi}]{khashabi2020unifiedqa}
Daniel Khashabi, Tushar Khot, Ashish Sabharwal, Oyvind Tafjord, Peter Clark,
  and Hannaneh Hajishirzi. 2020.
\newblock Unifiedqa: Crossing format boundaries with a single qa system.
\newblock \emph{arXiv preprint arXiv:2005.00700}.

\bibitem[{Liu and Singh(2004)}]{liu2004conceptnet}
Hugo Liu and Push Singh. 2004.
\newblock Conceptnet—a practical commonsense reasoning tool-kit.
\newblock \emph{BT technology journal}, 22(4):211--226.

\bibitem[{Liu et~al.(2019)Liu, Ott, Goyal, Du, Joshi, Chen, Levy, Lewis,
  Zettlemoyer, and Stoyanov}]{liu2019roberta}
Yinhan Liu, Myle Ott, Naman Goyal, Jingfei Du, Mandar Joshi, Danqi Chen, Omer
  Levy, Mike Lewis, Luke Zettlemoyer, and Veselin Stoyanov. 2019.
\newblock Roberta: A robustly optimized bert pretraining approach.
\newblock \emph{arXiv preprint arXiv:1907.11692}.

\bibitem[{Mitra et~al.(2019)Mitra, Banerjee, Pal, Mishra, and
  Baral}]{mitraadditional}
Arindam Mitra, Pratyay Banerjee, Kuntal Pal, Swaroop Mishra, and Chitta Baral.
  2019.
\newblock How additional knowledge can improve natural language commonsense
  question answering?
\newblock \emph{arXiv preprint arXiv:1909.08855}.

\bibitem[{Radford et~al.(2018)Radford, Narasimhan, Salimans, and
  Sutskever}]{radford2018improving}
Alec Radford, Karthik Narasimhan, Tim Salimans, and Ilya Sutskever. 2018.
\newblock Improving language understanding by generative pre-training.

\bibitem[{Radford et~al.(2019)Radford, Wu, Child, Luan, Amodei, and
  Sutskever}]{radford2019language}
Alec Radford, Jeffrey Wu, Rewon Child, David Luan, Dario Amodei, and Ilya
  Sutskever. 2019.
\newblock Language models are unsupervised multitask learners.

\bibitem[{Raffel et~al.(2019)Raffel, Shazeer, Roberts, Lee, Narang, Matena,
  Zhou, Li, and Liu}]{raffel2019exploring}
Colin Raffel, Noam Shazeer, Adam Roberts, Katherine Lee, Sharan Narang, Michael
  Matena, Yanqi Zhou, Wei Li, and Peter~J Liu. 2019.
\newblock Exploring the limits of transfer learning with a unified text-to-text
  transformer.
\newblock \emph{arXiv preprint arXiv:1910.10683}.

\bibitem[{Rajani et~al.(2019)Rajani, McCann, Xiong, and
  Socher}]{rajani2019explain}
Nazneen~Fatema Rajani, Bryan McCann, Caiming Xiong, and Richard Socher. 2019.
\newblock Explain yourself! leveraging language models for commonsense
  reasoning.
\newblock In \emph{Proceedings of the 57th Annual Meeting of the Association
  for Computational Linguistics}, pages 4932--4942.

\bibitem[{Sakaguchi et~al.(2019)Sakaguchi, Bras, Bhagavatula, and
  Choi}]{sakaguchi2019winogrande}
Keisuke Sakaguchi, Ronan~Le Bras, Chandra Bhagavatula, and Yejin Choi. 2019.
\newblock Winogrande: An adversarial winograd schema challenge at scale.
\newblock In \emph{Proceedings of the AAAI Conference on Artificial
  Intelligence}, pages 8732--8740.

\bibitem[{Sap et~al.(2019{\natexlab{a}})Sap, Le~Bras, Allaway, Bhagavatula,
  Lourie, Rashkin, Roof, Smith, and Choi}]{sap2019atomic}
Maarten Sap, Ronan Le~Bras, Emily Allaway, Chandra Bhagavatula, Nicholas
  Lourie, Hannah Rashkin, Brendan Roof, Noah~A Smith, and Yejin Choi.
  2019{\natexlab{a}}.
\newblock Atomic: An atlas of machine commonsense for if-then reasoning.
\newblock In \emph{Proceedings of the AAAI Conference on Artificial
  Intelligence}, volume~33, pages 3027--3035.

\bibitem[{Sap et~al.(2019{\natexlab{b}})Sap, Rashkin, Chen, Le~Bras, and
  Choi}]{sap2019socialiqa}
Maarten Sap, Hannah Rashkin, Derek Chen, Ronan Le~Bras, and Yejin Choi.
  2019{\natexlab{b}}.
\newblock Social iqa: Commonsense reasoning about social interactions.
\newblock In \emph{Proceedings of the 2019 Conference on Empirical Methods in
  Natural Language Processing and the 9th International Joint Conference on
  Natural Language Processing (EMNLP-IJCNLP)}, pages 4453--4463.

\bibitem[{Seo et~al.(2017)Seo, Kembhavi, Farhadi, and
  Hajishirzi}]{Seo2017Bidirectional}
Minjoon Seo, Aniruddha Kembhavi, Ali Farhadi, and Hannaneh Hajishirzi. 2017.
\newblock Bidirectional attention flow for machine comprehension.
\newblock In \emph{International Conference on Learning Representations}.

\bibitem[{Shwartz et~al.(2020)Shwartz, West, Bras, Bhagavatula, and
  Choi}]{shwartz2020unsupervised}
Vered Shwartz, Peter West, Ronan~Le Bras, Chandra Bhagavatula, and Yejin Choi.
  2020.
\newblock Unsupervised commonsense question answering with self-talk.
\newblock \emph{arXiv preprint arXiv:2004.05483}.

\bibitem[{Speer et~al.(2017)Speer, Chin, and Havasi}]{speer2017conceptnet}
Robyn Speer, Joshua Chin, and Catherine Havasi. 2017.
\newblock Conceptnet 5.5: An open multilingual graph of general knowledge.
\newblock In \emph{Proceedings of the AAAI Conference on Artificial
  Intelligence}.

\bibitem[{Talmor et~al.(2019)Talmor, Herzig, Lourie, and
  Berant}]{talmor2019commonsenseqa}
Alon Talmor, Jonathan Herzig, Nicholas Lourie, and Jonathan Berant. 2019.
\newblock {C}ommonsense{QA}: A question answering challenge targeting
  commonsense knowledge.
\newblock In \emph{Proceedings of the 2019 Conference of the North {A}merican
  Chapter of the Association for Computational Linguistics (NAACL)}, pages
  4149--4158.

\bibitem[{Wang et~al.(2018)Wang, Sun, Zhao, Shen, and
  Liu}]{Wang2018YuanfudaoAS}
Liang Wang, Meng Sun, Wei Zhao, Kewei Shen, and Jingming Liu. 2018.
\newblock Yuanfudao at semeval-2018 task 11: Three-way attention and relational
  knowledge for commonsense machine comprehension.
\newblock In \emph{Proceedings of The 12th International Workshop on Semantic
  Evaluation}, pages 758--762.

\bibitem[{Zellers et~al.(2019)Zellers, Holtzman, Bisk, Farhadi, and
  Choi}]{zellers-etal-2019-hellaswa}
Rowan Zellers, Ari Holtzman, Yonatan Bisk, Ali Farhadi, and Yejin Choi. 2019.
\newblock {H}ella{S}wag: Can a machine really finish your sentence?
\newblock In \emph{Proceedings of the 57th Annual Meeting of the Association
  for Computational Linguistics}, pages 4791--4800.

\bibitem[{Zhou et~al.(2018)Zhou, Young, Huang, Zhao, Xu, and
  Zhu}]{zhou2018commonsense}
Hao Zhou, Tom Young, Minlie Huang, Haizhou Zhao, Jingfang Xu, and Xiaoyan Zhu.
  2018.
\newblock Commonsense knowledge aware conversation generation with graph
  attention.
\newblock In \emph{IJCAI}.

\bibitem[{Zhu et~al.(2018)Zhu, Wei, Qin, and Liu}]{zhu2018hierarchical}
Haichao Zhu, Furu Wei, Bing Qin, and Ting Liu. 2018.
\newblock Hierarchical attention flow for multiple-choice reading
  comprehension.
\newblock In \emph{Proceedings of the AAAI Conference on Artificial
  Intelligence}.

\end{thebibliography}
\appendix

\end{document}